\documentclass{article} 
\usepackage[a4paper,headheight=20pt,top=25mm,bottom=25mm,left=35mm,right=35mm]{geometry}

\usepackage{arxiv}

\usepackage[T1]{fontenc}    
\usepackage{hyperref}       
\usepackage{amsfonts}       
\usepackage{fancyhdr}       
\usepackage{graphicx}       
\usepackage{amsmath}
\usepackage{amsthm}
\usepackage{hyphenat}
\usepackage{subfigure}
\usepackage{url}
\usepackage{algorithm}
\usepackage{algpseudocode}

\graphicspath{{img/}}

\title{Reinforcement Learning Your Way: Agent Characterization through Policy Regularization}

\author{Charl Maree
\thanks{Second affiliation: Chief Technology Office, Sparebank 1 SR-Bank, Stavanger, Norway.} 
\ \& Christian Omlin \\ Center for Artificial Intelligence Research\\ University of Agder\\ Grimstad, Norway \\ 
\texttt{\{charl.maree,christian.omlin\}@uia.no} \\ }

\begin{document}
\maketitle
\begin{abstract}
The increased complexity of state-of-the-art reinforcement learning (RL) algorithms have resulted in an opacity that inhibits explainability and understanding. This has led to the development of several post-hoc explainability methods that aim to extract information from learned policies thus aiding explainability. These methods rely on empirical observations of the policy and thus aim to generalize a characterization of agents' behaviour. In this study, we have instead developed a method to \emph{imbue} a characteristic behaviour into agents' policies through regularization of their objective functions. Our method guides the agents' behaviour during learning which results in an intrinsic characterization; it connects the learning process with model explanation. We provide a formal argument and empirical evidence for the viability of our method. In future work, we intend to employ it to develop agents that optimize individual financial customers' investment portfolios based on their spending personalities.
\end{abstract}

\section{Introduction}
Recent advances in reinforcement learning (RL) have increased complexity which, especially for deep RL, has brought forth challenges related to explainability \cite{heuillet2021a}. The opacity of state-of-the-art RL algorithms has led to model developers having a limited understanding of their agents' policies and no influence over learned strategies \cite{garcia2015}. While concerns surrounding explainability have been noted for AI in general, it is only more recently that attempts have been made to explain RL systems \cite{arrieta2020, heuillet2021a, wells21, gupta21}. These attempts have resulted in a wide suite of methods requiring various degrees of expert knowledge, either about the state-action domain or about the specific RL algorithm. Further, they typically rely on post-hoc analysis of learned policies which give only observational assurances of agents' behaviour. We instead propose an \emph{intrinsic} method of regularizing agents' actions based on a given prior. While current methods for RL regularization aim to improve \emph{training performance} - e.g., by maximizing the entropy of the action distribution \cite{haarnoja2017}, or by minimising the distance to a prior sub-optimal state-action distribution \cite{galashov2018} - our aim is the \emph{characterization} of our agents' behaviours. We also extend the current regularization techniques to accommodate multi-agent systems which allows intrinsic characterization of individual agents. We provide a formal argument for the rationale of our method and demonstrate its efficacy in a toy problem where agents learn to navigate to a destination on a grid by performing, e.g., only right turns (under the premise that right turns are considered safer than left turns \cite{lu2001}). There are several useful applications beyond this toy problem, such as financial advice based on personal goals. 

\section{Background and Related Work}
\subsection{Agent Characterization}
There have been several approaches to characterizing RL agents, with most - if not all - employing some form of post-hoc evaluation technique. Some notable examples are:
\begin{description}
    \item[Probabilistic argumentation \cite{riveret19}] in which a human expert creates an \sloppy{`argumentation graph'} with a set of arguments and sub-arguments; sub-arguments attack or support main arguments which attack or support discrete actions. Sub-arguments are labelled as `ON' or `OFF' depending on the state observation for each time-step. Main arguments are labelled as `IN', `OUT', or `UNDECIDED' in the following RL setting: \emph{states} are the union of the argumentation graph and the learned policy, \emph{actions} are the probabilistic `attitudes' towards given arguments, and \emph{rewards} are based on whether an argument attacks or supports an action. The learned `attitudes' towards certain arguments are used to characterize agents' behaviour.
    \item[Structural causal modelling (SCM) \cite{madumal2019a}] learns causal relationships between states and actions through `action influence graphs' that trace all possible paths from a given initial state to a set of terminal states, via all possible actions in each intermediate state. The learned policy then identifies a causal chain as the single path in the action influence graph that connects the initial state to the relevant terminal state. The explanation is the vector of rewards along the causal chain. Counter-explanations are a set of comparative reward vectors along chains originating from counter-actions in the initial state. Characterizations are made based on causal and counterfactual reasons for agents' choices of action.
    \item[Reward decomposition \cite{vanseijen17, juozapaitis2019a}] decomposes the reward into a vector of intelligible reward classes using expert knowledge. Agent characterization is done by evaluation of the reward vector for each action post training. 
    \item[Hierarchical reinforcement learning HRL \cite{beyret2019a, marzari21}] divides agents' tasks into sub-tasks to be learned by different agents. This simplifies the problem to be solved by each agent, making their behaviour easier to interpret and thereby making them easier to characterize. 
    \item[Introspection (interesting elements) \cite{sequeira2019a}] is a statistical post-hoc analysis of the policy. It considers elements such as the frequency of visits to states, the estimated values of states and state-action pairs, state-transition probabilities, how much of the state space is visited, etc. Interesting statistical properties from this analysis are used to characterize the policy.
\end{description}

\subsection{Multi-Agent Reinforcement Learning and Policy Regularization}
We consider the multi-agent setting of partially observable Markov decision processes (POMDPs) \cite{littman94}: for $N$ agents, let $\mathcal{S}$ be a set of states, $\mathcal{A}_i$ a set of actions, and $\mathcal{O}_i$ a set of incomplete state observations where $i \in [1,..,N]$ and $\mathcal{S} \mapsto \mathcal{O}_i$. Agents select actions according to individual policies $\pi_{\theta_i}(\mathcal{O}_i) \mapsto \mathcal{A}_i$ and receive rewards according to individual reward functions $r_i(\mathcal{S},\mathcal{A}_i) \mapsto \mathbb{R}$, where $\theta_i$ is the set of parameters governing agent $i$'s policy. Finally, agents aim to maximize their total discounted rewards: 
\begin{equation*} \label{eqn:discountedrewards}
    R_i(o,a) = \sum_{t=0}^{T} \gamma r_{i}(o_t,a_t)
\end{equation*}
where $T$ is the time horizon and $\gamma \in [0,1]$ is a discount factor. For single-agent systems, the deep deterministic policy gradient algorithm (DDPG) defines the gradient of the objective $J(\theta) = \mathbb{E}_{s \sim p^\mu}[R(s,a)]$ as \cite{lillicrap15}:
\begin{equation} \label{eqn:single-obj}
    \nabla_{\theta} J(\theta) = \mathbb{E}_{s \sim \mathcal{D}} \left[ \nabla_{\theta} \mu_{\theta} (a \vert s) \nabla_a Q^{\mu_\theta} (s,a) \vert_{a=\mu_{\theta}(s)} \right]
\end{equation}
where $p^\mu$ is the state distribution, $\mathcal{D}$ is an experience replay buffer storing observed state transition tuples $(s,a,r,s')$, and $Q^{\mu_\theta} (s,a)$ is a state-action value function where actions are selected according to a policy $\mu_\theta(\mathcal{S}) \mapsto \mathcal{A}$. In DDPG, the policy $\mu$ - also called the \emph{actor} - and the value function $Q$ - also called the \emph{critic} - are modelled by deep neural networks. Equation~(\ref{eqn:single-obj}) is extended to a multi-agent setting; the multi-agent deep deterministic policy gradient algorithm (MADDPG) learns individual sets of parameters for each agent $\theta_i$ \cite{lowe2017}:
\begin{equation} \label{eqn:multi-obj}
    \nabla_{\theta_i} J(\theta_i) = \mathbb{E}_{o,a \sim \mathcal{D}} \left[ \nabla_{\theta_i} \mu_{\theta_i} (a_i \vert o_i) \nabla_{a_i} Q^{\mu_{\theta_i}} \left( o_i,a_1,...,a_N \right) \vert_{a_i=\mu_{\theta_i}(o_i)} \right]
\end{equation}
where $o_i \in \mathcal{O}_i$ and the experience replay buffer $\mathcal{D}$ contains tuples $(o_i,a_i,r_i,o'_i), \ i \in [1,..,N]$.

In this work, we further extend MADDPG by adding a regularization term to the actors' objective functions, thus encouraging them to mimic the behaviours specified by simple predefined prior policies. There have been several approaches to regularizing RL algorithms, mostly for the purpose of improved generalization or training performance. In \cite{galashov2018}, the authors defined an objective function with a regularization term related to the statistical difference between the current policy and a predefined prior:
\begin{equation} \label{eqn:KLreg}
    J(\theta) = \mathbb{E}_{s,a \sim \mathcal{D}} \left[ R\left(s,a\right) - \alpha D_{KL} \left( \pi_\theta \left(s, a \right) \| \pi_0 \left(s,a \right) \right) \right]
\end{equation}
where $\alpha$ is a hyperparameter scaling the relative contribution of the regularization term - the Kullback-Leibler (KL) divergence ($D_{KL}$) - and $\pi_{0}$ is the prior policy which the agent attempts to mimic while maximising the reward. The KL divergence is a statistical measure of the difference between two probability distributions, formally:
\begin{equation*} \label{eqn:D_KL}
    D_{KL}(P \| Q) = \sum_{x \in X} P(x) \log{\frac{P(x)}{Q(x)}}
\end{equation*}
where $P$ and $Q$ are discrete probability distributions on the same probability space $X$. The stated objective of KL regularization is increased learning performance by penalising policies that stray too far from the prior. The KL divergence is often also called the \emph{relative entropy}, with KL-regularized RL being the generalization of entropy-regularized RL \cite{ziebart2010}; specifically if $\pi_0$ is the uniform distribution, Equation \ref{eqn:KLreg} reduces to, up to a constant, the objective function for entropy-regulated RL as described in \cite{haarnoja2017}:
\begin{equation} \label{eqn:EntropyReg}
    J(\theta) = \mathbb{E}_{s,a \sim \mathcal{D}} \left[ R(s,a) + \alpha H \left[ \pi_\theta(s,a) \right] \right]
\end{equation}
where $H(\pi)=P(\pi)\log(P(\pi))$ is the statistical entropy of the policy. The goal of entropy-regularized RL is to encourage exploration by maximising the policy's entropy and is used as standard in certain state-of-the-art RL algorithms, such as soft actor-critic (SAC) \cite{haarnoja2017}. Other notable regularization methods include \emph{control regularization} where, during learning, the action of the actor is weighted with an action from a sub-optimal prior: $\mu_k = \frac{1}{1+\lambda} \mu_\theta + \frac{\lambda}{1+\lambda} \mu_{prior}$ and \emph{temporal difference regularization} which adds a penalty for large differences in the Q-values of successive states: $J(\theta, \eta) = \mathbb{E}_{s,a,s' \sim \mathcal{D}} \left[ R(s,a) -\eta \delta_Q(s,a,s') \right]$ where $\delta_Q(s,a,s') = \left[ R(s,a) + \gamma Q(s',a') - Q(s,a)) \right]^2$ \cite{cheng2019, parisi2019}.

While our algorithm is based on regularization of the objective function, it could be argued that it shares similar goals as those of algorithms based on \emph{constrained RL}, namely the intrinsic manipulation of agents' policies towards given objectives. One example of constrained RL is \cite{Miryoosefi2019} which finds a policy whose long-term measurements lie within a set of constraints by penalising the reward function with the Euclidean distance between the state and a given set of restrictions, e.g., an agent's location relative to obstacles on a map. Another example is  \cite{chow2015} which penalises the value function with the accumulated cost of a series of actions, thus avoiding certain state-action situations. However, where constrained RL attempts to \emph{avoid} certain conditions - usually through a penalty based on expert knowledge of the state - regularized RL aims to \emph{promote} desired behaviours, such as choosing default actions during training or maximizing exploration by maximising action entropy. The advantage of our system is that it does not require expert knowledge of the state-action space to construct constraints; our regularization term is independent of the state which allows agents to learn simple behavioural patterns, thus improving the interpretability of their characterization.
\section{Methodology}
We regulate our agents based on a state-independent prior to maximize rewards while adhering to simple, predefined rules. In a toy problem, we demonstrate that agents learn to find a destination on a map by taking only right turns. Intuitively, we supply the probability distribution of three actions - \emph{left}, \emph{straight}, and \emph{right} - as a regularization term in the objective function, meaning the agents aim to mimic this given probability distribution while maximising rewards. Such an agent can thus be characterized as an agent that prefers, e.g., right turns over left turns. As opposed to post-hoc characterization, ours is an intrinsic method that inserts a desirable characteristic into an agent's behaviour \emph{during} learning. 

\subsection{Action Regularization}
We modify the objective function in Equation~(\ref{eqn:EntropyReg}) and replace the regularization term \sloppy{$H[\pi_\theta(s,a)] = P(\pi) \log (P(\pi))$} with the mean squared error of the expected action and a specified prior $\pi_0$:
\begin{align} \label{eqn:obj_reg}
    J(\theta) &= \mathbb{E}_{s,a \sim \mathcal{D}} \left[ R(s,a) \right] - \lambda L \\
    L &= \frac{1}{M} \sum_{j=0}^{M} \Big[ \mathbb{E}_{a \sim \pi_\theta} \left[ a_j \right] - (a_{j} \vert \pi_0(a)) \Big]^2
\end{align}
where $\lambda$ is a hyperparameter that scales the relative contribution of the regularization term $L$, $a_j$ is the $j^{th}$ action in a vector of $M$ actions, $\pi_\theta$ is the current policy, and $\pi_0$ is the specified prior distribution of actions which the agent aims to mimic while maximising the reward. Note that $\pi_0(a)$ is \emph{independent} of the state and $(a_{j} \vert \pi_0(a))$ is therefore constant across all observations and time-steps. This is an important distinction from previous work, and results in a prior that is simpler to construct and a characterization that can be interpreted by non-experts. Since this is a special case of Equation~(\ref{eqn:EntropyReg}), it follows from the derivation given in \cite{haarnoja2017}. 

We continue by extending our objective function to support a multi-agent setting. From Equation~(\ref{eqn:obj_reg}) and following the derivation in \cite{lowe2017}, we derive a multi-agent objective function with $i \in [1,N]$ where $N$ is the number of agents:
\begin{equation} \label{eqn:ma_j}
    J(\theta_i) = \mathbb{E}_{o_i,a_i \sim \mathcal{D}} \left[ R_i(o_i,a_i) \right] - \lambda \frac{1}{M_i} \sum_{j=0}^{M_i} \left[ \mathbb{E}_{a \sim \pi_{\theta_i}(o_i,a)}(a_j) - (a_{j} \vert \pi_{0_i}(a)) \right]^2
\end{equation}
Further, in accordance with the MADDPG algorithm, we model actions and rewards with actors and critics, respectively \cite{lowe2017}:
\begin{align} 
    \mathbb{E}_{a_i \sim \pi_{\theta_i}(o_i,a_i)} \left[ a_i \right] &= \mu_{\theta_i}(o_i) \label{eqn:mu} \\
    \mathbb{E}_{o_i,a_i \sim \mathcal{D}} \left[ R_i(o_i,a_i) \right] &= Q_{\theta_i}(o_i,\mu_{\theta_1}(o_1),\dots,\mu_{\theta_N}(o_N)) \label{eqn:q}
\end{align}
Through simple substitution of Equations (\ref{eqn:mu}) and (\ref{eqn:q}) into Equation~(\ref{eqn:ma_j}) we formulate our multi-agent regularized objective function:
\begin{align} \label{eqn:multi_agent}
    J(\theta_i) &= Q_{\theta_i}(o_i,\mu_{\theta_1}(o_1),\dots,\mu_{\theta_N}(o_N)) - \lambda L_i \\
    L_i &= \frac{1}{M_i} \sum_{j=0}^{M_i} \Big[ \mu_{\theta_i}(o_i)_j - (a_{j} \vert \pi_{0_i}(a)) \Big]^2
\end{align}

Algorithm \ref{alg:rmaddpg} optimizes the policies of multiple agents given individual regularization constraints $\pi_{0_i}$.

\begin{algorithm}
\caption{Action-regularized MADDPG algorithm} \label{alg:rmaddpg}
\begin{algorithmic}
\State Set the number of agents $N \in \mathbb{N}$
\For{i in 1, N} \Comment{For each agent}
    \State Initialize actor network $\mu_{\theta_{\mu,i}}$ with random parameters $\theta_{\mu,i}$
    \State Initialize critic network $Q_{\theta_{Q,i}}$ with random parameters $\theta_{Q,i}$
    \State Initialize target actor network $\mu'_{\theta_{\mu',i}}$ with parameters $\theta_{\mu',i} \gets \theta_{\mu,i}$
    \State Initialize target critic network $Q'_{\theta_{Q',i}}$ with parameters $\theta_{Q',i} \gets \theta_{Q,i}$
    \State Set the desired prior action distribution $\pi_{0_i}$
    \State Set the number of actions $M_i \gets |\pi_{0_i}|$
\EndFor
\State Initialize replay buffer $\mathcal{D}$
\State Set regularization weight $\lambda$
\For{e = 1, Episodes}
    \State Initialise random function $F(e) \sim N(0,\sigma_e)$ for exploration
    \State Reset environment and get the state observation $s_1 \mapsto o_{[1..N]}$
    \State $t \gets 1$, $Done \gets False$
    \While{not Done}
        \For{i in 1, N} \Comment{For each agent}
            \State Select action with exploration $a_{i,t} \gets \mu_{\theta_{\mu,i}}(o_i) + F(e)$
        \EndFor
        \State Apply compounded action $a_t$
        \State Retrieve rewards $r_{[1..N],t}$ and observations $s_{t}' \mapsto o'_{[1..N],t}$
        \State Store transition tuple $\mathcal{T} = \left( o_t, a_t, r_t, o'_t \right)$ to replay buffer: $\mathcal{D} \gets \mathcal{D} \cup \mathcal{T}$
        \State $t \gets t+1$
        \If{(end of episode)}
            \State $Done \gets True$
        \EndIf
    \EndWhile
    \State Sample a random batch from the replay buffer $\mathcal{B} \subset \mathcal{D}$
    \For{i in 1, N} \Comment{For each agent}
        \State $\widehat{Q_i} \gets r_{\mathcal{B},i} + \gamma Q'_i \left( o'_{\mathcal{B},i}, \mu'_1\left(o'_{\mathcal{B},1} \right) , \dots, \mu'_N\left(o'_{\mathcal{B},N} \right) \right)$
        \State Update critic parameters $\theta_{Q,i}$ by minimising the loss: 
        \begin{equation*}
            \mathcal{L}(\theta_{Q,i}) = \frac{1}{|\mathcal{B}|} \sum_{\mathcal{B}} \left(Q_{\theta_{Q,i}} \left( o_\mathcal{B},a_{\mathcal{B},1},\dots,a_{\mathcal{B},N} \right) - \widehat{Q_i} \right)^2
        \end{equation*}
        \State Update the actor parameters $\theta_{\mu,i}$ by minimising the loss: \Comment{From Equation \ref{eqn:multi_agent}}
        \begin{align*}
            \widehat{R_i} &= \overline{Q_i \left( o_{\mathcal{B},i}, \mu_1 (o_{\mathcal{B},1}),\dots,\mu_N (o_{\mathcal{B},N}) \right)} \\
            \mathcal{L}(\theta_{\mu,i}) &= - \widehat{R_i} + \lambda \frac{1}{M_i} \sum_{j=1}^{M_i} \left[ \overline{ \mu_i (o_{\mathcal{B},i})_{j} } - \left( a_j \vert \pi_{0_i} \right) \right]^2
        \end{align*}
        \State Update target network parameters:
        \begin{align*}
            \theta_{\mu',i} &\gets \tau \theta_{\mu,i} + \left( 1-\tau \right) \theta_{\mu',i} \\
        \theta_{Q',i} &\gets \tau \theta_{Q,i} + \left( 1-\tau \right) \theta_{Q',i}
        \end{align*}
    \EndFor    
\EndFor
\end{algorithmic}
\end{algorithm}

\section{Experiments}
\subsection{Empirical Setup}
We created a toy problem in which one or more agents navigate a $6 \times 6$ grid through a set of three actions: $\mathcal{A}_{1} =$ \emph{turn left}, $\mathcal{A}_{2} =$ \emph{go straight}, and $\mathcal{A}_{3} =$ \emph{turn right}. Every new episode randomly placed a set of destinations in the grid $D_i, \, i \in [1,N]$ - one for each of $N \geq 1$ agents - with initial agent locations $L_{i,0} = (3,0)$. Rewards were the agents' Euclidean distances from their destinations $R_{i,t} = \lVert D_{i} - L_{i,t} \rVert_2 $ where $L_{i,t}$ is the location of agent $i$ at time-step $t$. Finally, agents' observations were the two-dimensional distances to their destinations: $\mathcal{O}_{i,t} = D_{i} - L_{i,t}$. An episode was completed when either both agents had reached their destinations or a maximum of 50 time-steps had passed.

We ran two sets of experiments, one for a single-agent setting, and one for a dual-agent setting. We sized all networks in these two settings with two fully connected feed-forward layers; the single agent networks had 200 nodes in each layer, while the dual-agent networks had 700 nodes in each layer. Actor networks had a softmax activation layer, while the critic networks remained unactivated. Our training runs consisted of 3~000 iterations and we tuned the hyperparameters using a simple one-at-a-time parameter sweep. We used training batches of 256 time-steps and sized the reply buffers to hold 2~048 time-steps. In each iteration we collected 256 time-steps and ran two training epochs. We tuned the learning rates to 0.04 for the actors and 0.06 for the critics, the target network update parameters $\tau$ to 0.06, and the discount factors $\gamma$ to 0.95. We specified the regularization coefficient $\lambda = 2$, the regularization prior for the single-agent setting as $\pi_0 = [P(\mathcal{A}_{1}), P(\mathcal{A}_{2}), P(\mathcal{A}_{3})] = [0.0, 0.6, 0.4]$, and the regularization priors for the dual-agent setting as $\pi_{0,1} = [0.0, 0.6, 0.4]$ and $\pi_{0,2} = [0.4, 0.6, 0.0]$. This meant that the single agent was regularized to not take any left turns, while slightly favouring going straight above turning right. For the dual agents, agent 1 was to avoid left turns while agent 2 was to instead avoid right turns; we did this to demonstrate the characterization of the agents as preferring either left or right turns while navigating to their destinations.

\subsection{Results}
\begin{figure}[h]
    \centering
    \subfigure[The agent made a series of right turns to reach its destination on the left.]{
        \includegraphics[width=0.3\linewidth]{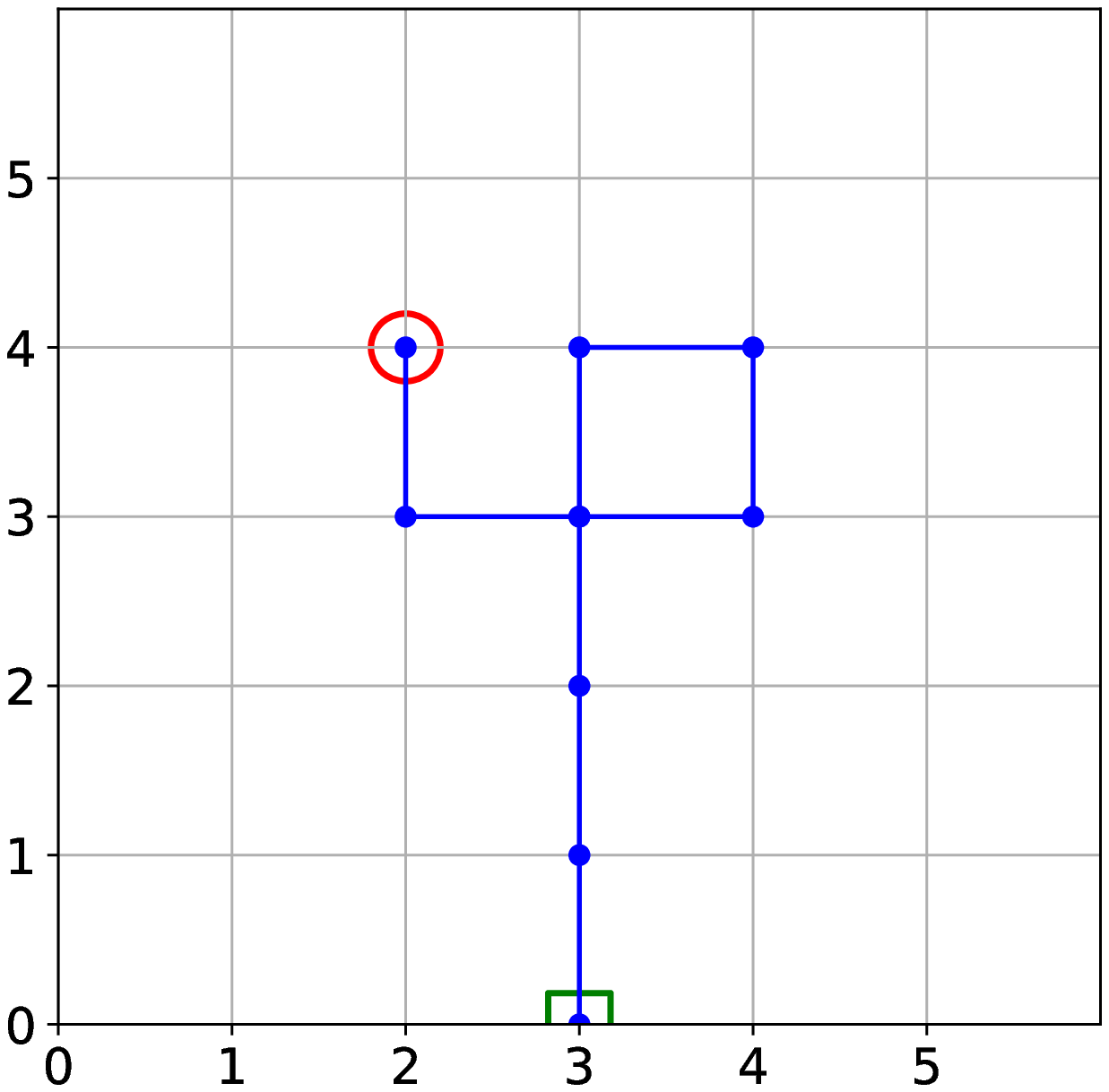}
    }\hfill
    \subfigure[The destination was straight ahead and the agent needed not turn.]{
        \includegraphics[width=0.3\linewidth]{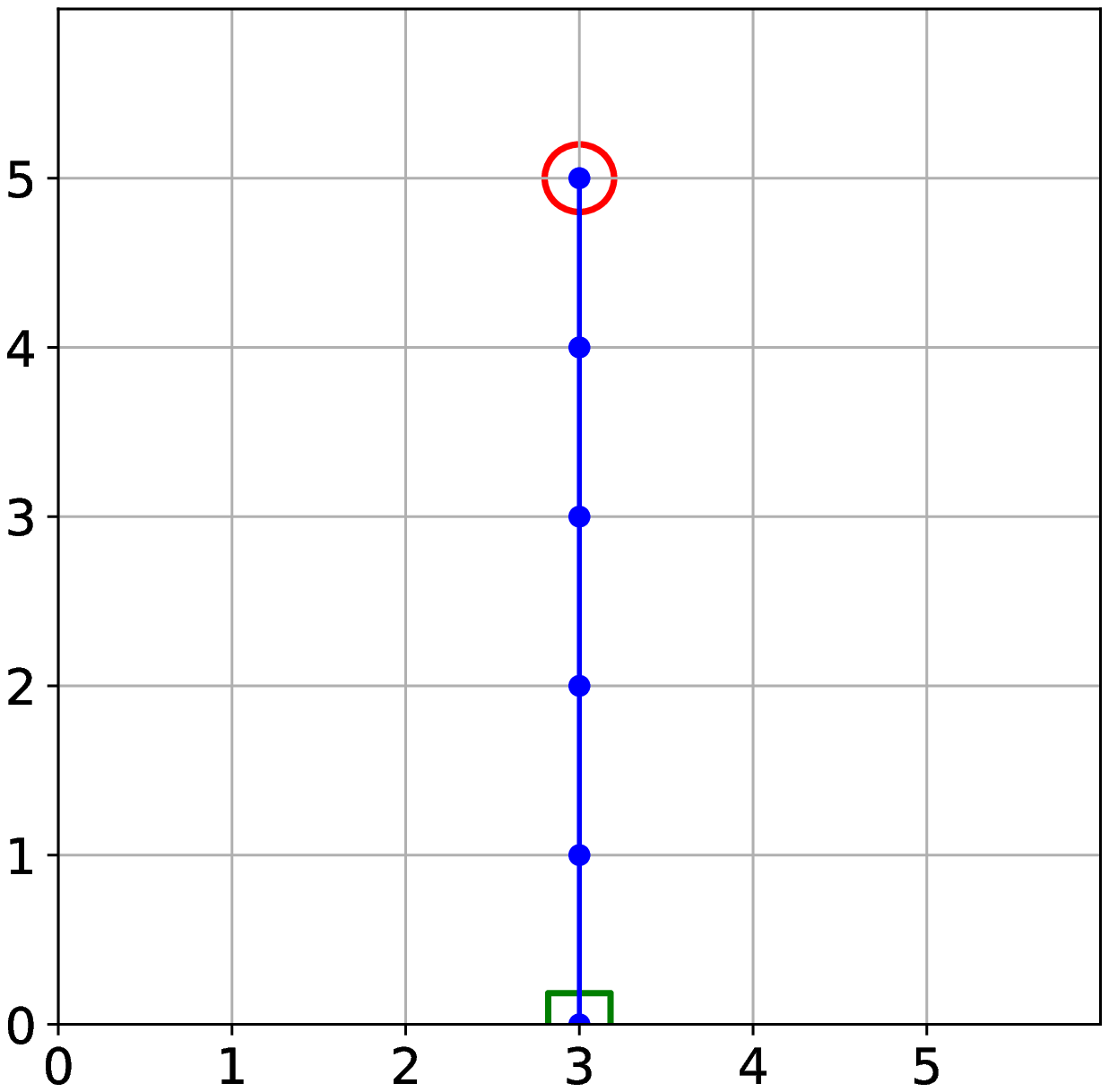}
    }\hfill
    \subfigure[The agent followed the shortest path and made a right turn.]{
        \includegraphics[width=0.3\linewidth]{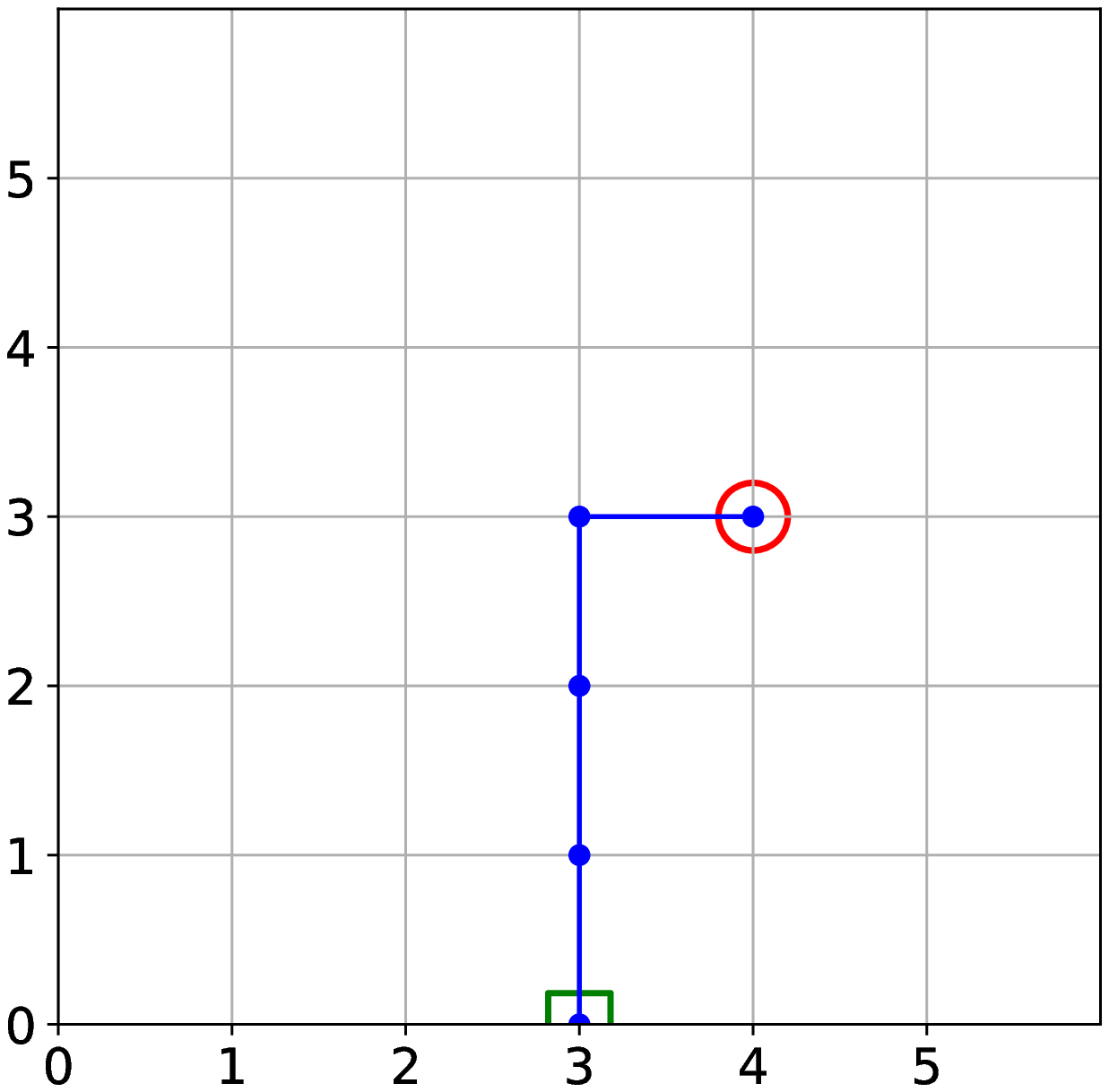}
    }
    \caption{Three trajectories of a single agent in the navigation problem. The starting locations are consistently (3,0) and the destinations are marked by red circles. During learning, the agent received a regularization prior $\pi_0 = [0.0, 0.6, 0.4]$ where the values in $\pi_0$ correspond to the probabilities of the actions \emph{turn left}, \emph{go straight}, and \emph{turn right} respectively.}
    \label{fig:sa_setting}
\end{figure}

In the single agent setting of our toy problem, we used our algorithm to encourage an agent to prefer right turns over left turns; we used a regularization prior $\pi_0 = [0.0, 0.6, 0.4]$ to regulate the probability of \emph{left actions} to $0.0$, \emph{straight actions} to $0.6$, and \emph{right actions} to $0.4$. Figure \ref{fig:sa_setting} shows three different trajectories that demonstrate such an agent's behaviour for destinations which lie either to the left, straight ahead, or to the right of the agent's starting location. As expected, the agent never turned left and always took the shortest route to its destination given its constraints.

\begin{figure}[!ht]
    \centering
    \subfigure[Both agents' destinations were on the left, but only the agent regularized to prefer left turns actually turned left. The other agent completed a series of right turns to reach its destination.]
    {
        \includegraphics[width=0.4\linewidth]{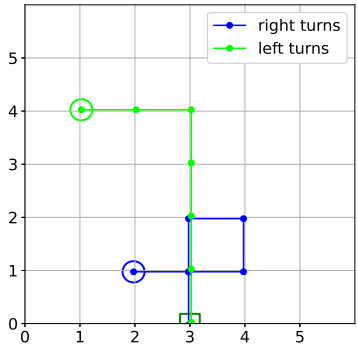}
    }\hfil
    \subfigure[Both agents' destinations were located such that they had to perform a series of turns in the direction according to their regularization priors $\pi_{0,i}$]
    {
        \includegraphics[width=0.4\linewidth]{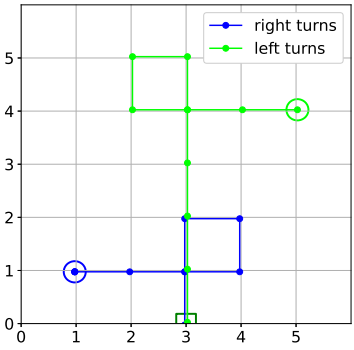}
    }\\
    \subfigure[Both agents' destinations were located such that they could perform their preferential turn - either to the left, or to the right - according to their regularization priors $\pi_{0,i}$]
    {
        \includegraphics[width=0.4\linewidth]{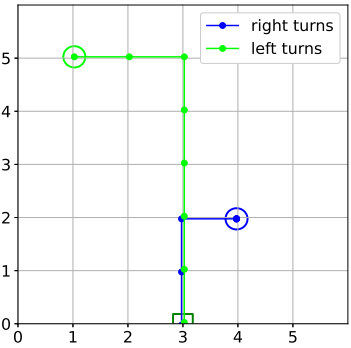}
    }\hfil
    \subfigure[One agent's destination was straight ahead and it needed not turn. Clearly, the regularization priors $\pi_{0,i}$ allowed for such strategies.]
    {
        \includegraphics[width=0.4\linewidth]{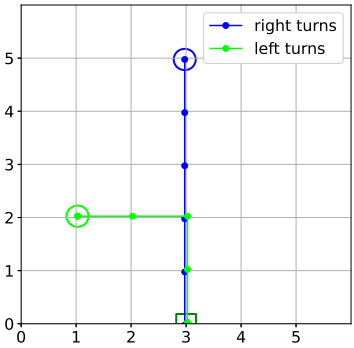}
    }
    \caption{Four sets of trajectories for a dual-agent environment in the navigation problem. The first agent - labelled 'right turns' - received a regularization prior $\pi_{0,1} = [0.0, 0.6, 0.4]$ while the second agent - labelled 'left-turns' received a regularization prior $\pi_{0,2} = [0.4, 0.6, 0.0]$ where the values in $\pi_{0,i}$ correspond to the probabilities of the actions \emph{turn left}, \emph{go straight}, and \emph{turn right}.}
    \label{fig:ma_seeting}
\end{figure}

Figure \ref{fig:ma_seeting} shows the same grid navigation problem, but this time for a multi-agent setting. Here, we used two agents with different regularization terms to constrain their actions to 1) right turns only and 2) left turns only; the first agent's regularization prior $\pi_{0,1} = [0.0, 0.6, 0.4]$ specified a probability of $0.0$ for the \emph{left action}, $0.6$ for the \emph{straight action}, and $0.4$ for the \emph{right action}, while the second agent's regularization prior $\pi_{0,2} = [0.4, 0.6, 0.0]$ specified a probability of $0.4$ for the \emph{left action}, $0.6$ for the \emph{straight action}, and $0.0$ for the \emph{right action}. Clearly, the two agents have learned different strategies in the navigation problem. In Figure \ref{fig:ma_seeting} it is clear that the two agents consistently took the shortest path to their respective destinations while adhering to their individual constraints. We therefore characterize them as agents that preferred to take right and left turns, respectively. Crucially, this is an \emph{intrinsic} property of the agents imposed by the regularization of the objective function. This separates our method of intrinsic characterization from post-hoc characterization techniques.

\begin{figure}[!hb]
    \centering
    \includegraphics[width=0.9\linewidth]{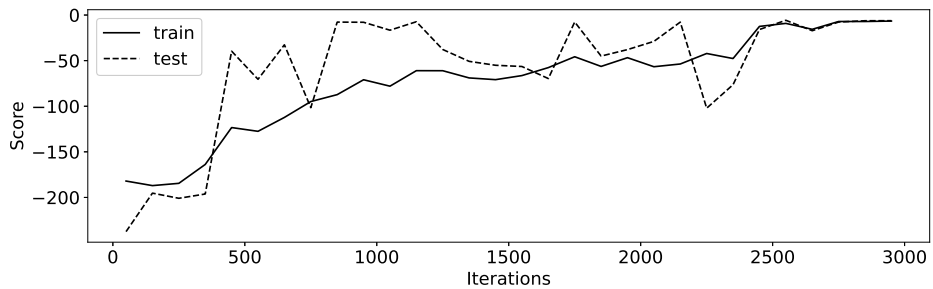}
    \caption{Training and testing returns for a typical training run. The learning process clearly followed a steady increase in returns.}
    \label{fig:returns}
\end{figure}

Finally, Figure \ref{fig:returns} shows a typical curve of training and testing returns across the 3~000 training iterations. The agents clearly demonstrate a good learning response with steadily increasing returns both in training and testing. 

\section{Conclusions and Direction for Future Work}
Our objective was the \emph{intrinsic} characterization of RL agents. To this end, we investigated and briefly summarized the relevant state-of-the-art in explainable RL and found that these methods have typically been relying on post-hoc evaluations of a learned policy. Policy regularization is a method that modifies a policy; however, it has typically been employed to enhance training performance which does not necessarily aid in policy characterization. We therefore adapted entropy regularization from maximizing the entropy in the policy to minimizing the mean squared difference between the expected action and a given prior. This encourages the agent to mimic a predefined behaviour while maximizing its reward during learning. Finally, we extended MADDPG with our regularization term. We provided a formal argument for the validity of our algorithm and empirically demonstrated its functioning in a toy problem. In this problem, we characterized two agents to follow different approaches when navigating to a destination in a grid; while one agent performed only right turns, the other performed only left turns. We conclude that our fundamentally sound algorithm was able to imbue specific characteristic behaviours into our agents' policies. In future work, we intend to use this algorithm to develop a set of financial advisors that will optimize individual customers' investment portfolios according to their individual spending personalities \cite{maree2021}. While maximising portfolio values, these agents may prefer, e.g., property investments over crypto currencies which are analogous to right turns and left turns in our toy problem.

\bibliographystyle{unsrt}
\bibliography{references}
\end{document}